\documentclass{article}

\usepackage[preprint, nonatbib]{neurips_2022}

\usepackage[utf8]{inputenc} % allow utf-8 input
\usepackage[T1]{fontenc}    % use 8-bit T1 fonts
\usepackage{hyperref}       % hyperlinks
\usepackage{url}            % simple URL typesetting
\usepackage{booktabs}       % professional-quality tables
\usepackage{amsfonts}       % blackboard math symbols
\usepackage{nicefrac}       % compact symbols for 1/2, etc.
\usepackage{microtype}      % microtypography
\usepackage{xcolor}         % colors

\usepackage[square, numbers]{natbib}
\usepackage[pdftex]{graphicx}
\usepackage{wrapfig}
\usepackage{amsmath}

\usepackage[ruled, lined, linesnumbered, commentsnumbered, noend]{algorithm2e}
\usepackage{soul}  %for code highlight
\DeclareRobustCommand{\hlgray}[1]{{\sethlcolor{lightgray}\hl{#1}}}

\title{Efficient Exploration using Model-Based Quality-Diversity with Gradients}

\author{%
    Bryan Lim\thanks{Equal Contribution} \\
    Imperial College London \\
    London, United Kingdom \\
    \texttt{bwl116@ic.ac.uk} \\
    \And
    Manon Flageat\footnotemark[1] \\
    Imperial College London \\
    London, United Kingdom \\
    \texttt{mf4618@ic.ac.uk} \\
    \And
    Antoine Cully \\
    Imperial College London \\
    London, United Kingdom \\
    \texttt{a.cully@imperial.ac.uk} \\
}
\usepackage{enumitem}

\usepackage{stmaryrd}
\usepackage{etoolbox}
\usepackage{multirow}
\begin{document}

\maketitle

\begin{abstract}
Exploration is a key challenge in Reinforcement Learning, especially in long-horizon, deceptive and sparse-reward environments. 
For such applications, population-based approaches have proven effective.
Methods such as Quality-Diversity deals with this by encouraging novel solutions and producing a diversity of behaviours.
However, these methods are driven by either undirected sampling (i.e. mutations) or use approximated gradients (i.e. Evolution Strategies) in the parameter space, which makes them highly sample-inefficient.
In this paper, we propose a model-based Quality-Diversity approach.
It extends existing QD methods to use gradients for efficient exploitation and leverage perturbations in imagination for efficient exploration.
Our approach optimizes all members of a population simultaneously to maintain both performance and diversity efficiently by leveraging the effectiveness of QD algorithms as good data generators to train deep models.
We demonstrate that it maintains the divergent search capabilities of population-based approaches on tasks with deceptive rewards while significantly improving their sample efficiency and quality of solutions.
\end{abstract}

\newcommand{\longname}[0]{Gradient and Dynamics Aware QD}
\newcommand{\name}[0]{GDA-QD}
\newcommand{\longnamedaqd}[0]{Dynamics-Aware QD}
\newcommand{\namedaqd}[0]{DA-QD-ext}
\newcommand{\sourcecode}[0]{TO-BE-RELEASED}

\newcommand{\expect}[1]{ \mbox{E} \left[ {#1} \right]}

% Notations used everywhere in the paper
\newcommand{\pop}[0]{\Theta}
\newcommand{\imagpop}[0]{\tilde{\Theta}}
\newcommand{\addbuffer}[0]{\mathcal{B}_{add}}
\newcommand{\batchsize}[0]{B}
\newcommand{\numimagiterations}[0]{N}
\newcommand{\iter}[0]{j}
\newcommand{\numiterations}[0]{J}
\newcommand{\popit}[0]{\Theta_j}
\newcommand{\imagpopit}[0]{\tilde{\Theta}_j}
\newcommand{\modeltrainperiod}[0]{J_{model}}
\newcommand{\maxmodelsteps}[0]{n_{steps}}

\section{Introduction}
\label{sec:intro}
\vspace{-2mm}
Reinforcement Learning (RL) has demonstrated tremendous abilities to learn challenging tasks across a range of applications~\citep{mnih2015human, silver2016mastering, akkaya2019solving}. 
However, they generally struggle with exploration as the agent can only gather data by interacting with the environment.
On the other hand, population based learning methods have shown to be very effective approaches ~\citep{jaderberg2017population, vinyals2019grandmaster, ecoffet2021first, wang2020enhanced}. 
In contrast to single agent learning, training a population of agents allow diverse behaviors and data to be collected.
This results in exploration that can better handle sparse and deceptive rewards~\cite{ecoffet2021first} as well as alleviate catastrophic forgetting~\citep{conti2018improving}.

An effective way to use the population of agents for exploration are novelty search methods~\citep{lehman2011evolving, conti2018improving} where the novelty of the behaviors of new agents is measured with respect to the population.
This novelty measure is then used in place of the conventional task reward similar to curiosity and intrinsic motivation approaches~\citep{oudeyer2007intrinsic, bellemare2016unifying, pathak2017curiosity}.
Quality-Diversity (QD)~\cite{pugh2016quality, cully2015robots, chatzilygeroudis2021quality} extends this but also optimizes all members of the population on the task reward while maintaining the diversity through novelty.
Beyond exploration, the creativity involved in finding various ways to solve a problem/task (i.e. the QD problem) is an interesting aspect of general intelligence that is also associated with adaptability. 
For instance, discovering diverse walking gaits can enable rapid adaptation to damage~\citep{cully2015robots}.

\begin{figure*}[t!]
\centering
    \includegraphics[width = 0.9\textwidth]{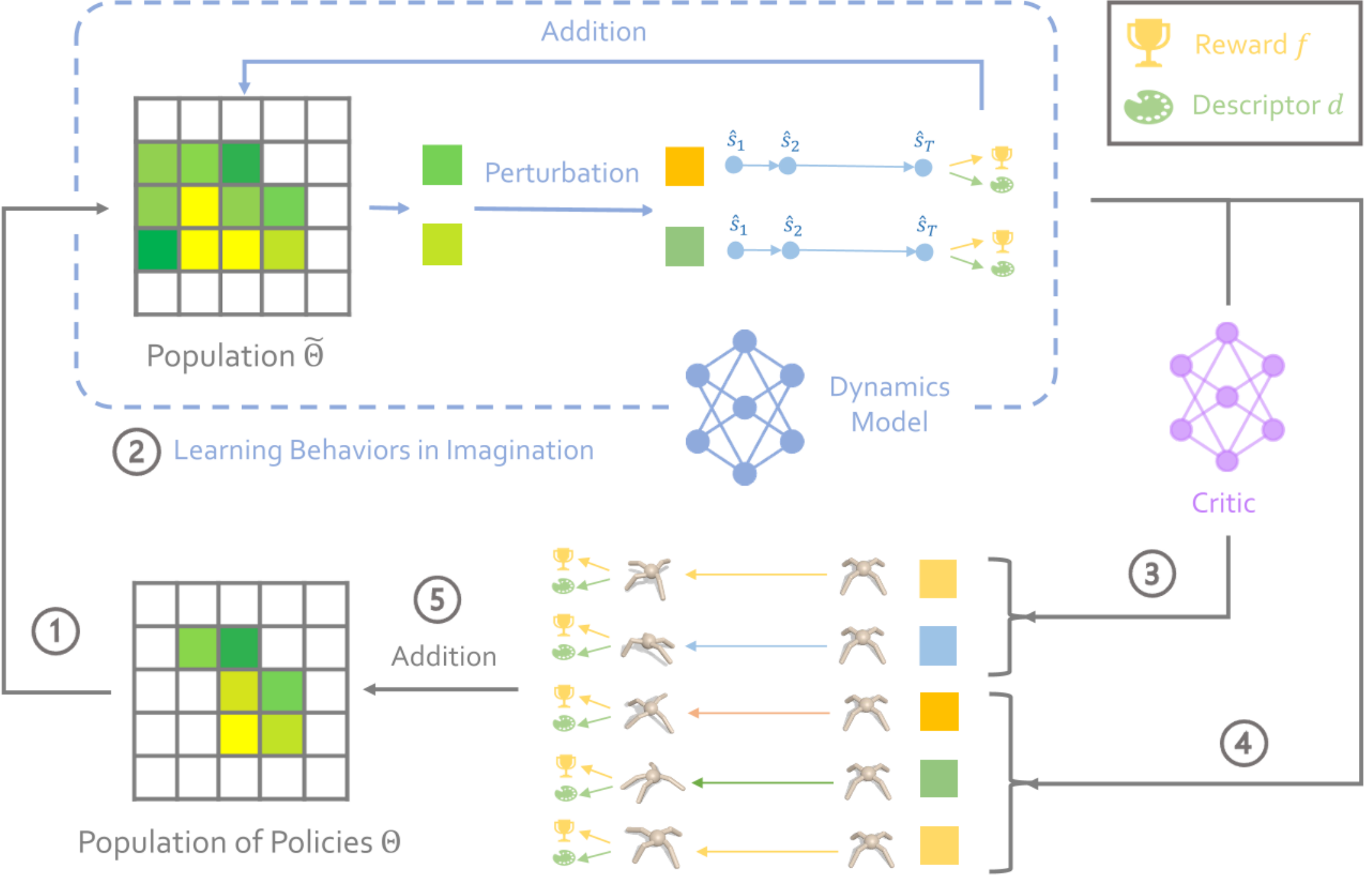}
    \caption{
    % \name{} optimizes all members of a population of policies simultaneously to maintain both performance and diversity. Its loop can be summarized as followed:
    The \name{} algorithm can be summarized as follows:
    (1) the current population $\pop{}$ is copied in $\imagpop{}$,
    (2) $\imagpop{}$ is used to perform multiple steps of QD optimization fully in imagination using the dynamics model,
    % (3) the transitions collected in imagination are used to train a critic network,
    % (4) the critic is used to apply policy-gradients updates to improve the policies in the population % most promising solutions in . 
    (3) the critic update sampled policies learned in imagination with policy-gradient updates; they are concatenated with
    (4) policies sampled from the resulting population learned in imagination $\imagpop{}$,
    (5) these concatenated batch of policies are evaluated in the environment and used to update the real population of policies for the next optimization loop; the transitions collected in the environment are then used to train the dynamics model and the critic.}
\label{fig:main_figure}
\end{figure*}

However, a drawback of conventional population based approaches is the large amounts of samples and evaluations required, usually in the order of millions.
Some methods that utilize Evolutionary Strategies (ES) and more recently MAP-Elites~\cite{mouret2015illuminating} (a common QD algorithm), sidestep this issue as they can parallelize and scale better with compute~\citep{salimans2017evolution, conti2018improving, lim2022accelerated} than their Deep RL counterparts, resulting in faster wall-clock times.
Despite this, they still come at a cost of many samples.
% fast - short amount of wall clock time as they scale well to distirbuted parallel computing systems
One of the main reasons for this lies in the underlying optimization operators. 
QD methods generally rely on undirected search methods such as objective-agnostic random perturbations~\citep{mouret2015illuminating, vassiliades2018iso} to favor creativity and exploration.
More directed search such as ES has also been used~\citep{colas2020scaling} but relies on a large number of such perturbations ($\sim$thousands) to approximate a single step of natural gradient to direct the improvement of solutions.

% 
% In this paper, we introduce a model-based Quality-Diversity approach which optimizes the entire population in imagination.
% Additionally, we also build on policy gradients~\cite{nilsson2021policy}

In this paper, we introduce an extended version of \longnamedaqd{} (\namedaqd{}) as well as \longname{} (\name{}), a new model-based QD method to perform sample-efficient exploration in RL.
\name{} optimizes an entire population of diverse policies through a QD  process in imagination using a learned dynamics model.
Additionally, \name{} augments the conventional QD optimization operators with policy gradient updates using a critic network to obtain a more performant population.
Beyond the effective exploration capabilities of QD methods, they are also excellent data generators.
We leverage this idea to harvest a diversity of transitions to train the dynamics model and the critic.
%Similarly, we use the QD process in imagination to train a critic network using the large diversity of imaginary transitions generated.
% \name{} then augments the conventional QD optimization operators with policy gradient information.
% The QD process is also used to harvest a diversity transitions, used to improve the policies in the population through gradients.
Thus, \name{} combine the powerful function-approximation capabilities of deep neural networks with the directed-search abilities of gradient-based learning and the creativity of population-based approaches. 
We demonstrate that it successfully outperforms both Deep RL and QD baselines in a hard-exploration task. 
\name{} exceeds the performance of baseline QD algorithms by $\sim 1.5$ times, and can reach the same results in $5$ times less samples. 
% It also finds global optimal policy performing $2$ times better than Deep RL baseline.

% two fold 
% firstly, the diversity of behaviors and transitions provide a good dataset to learn the dynamics/world model.
% Addoitionally, performing QD in imagination also procides a diversity of behaviours and transitions to train a critic.

\section{Preliminaries}
\vspace{-2mm}
\subsection{Reinforcement Learning}
\vspace{-2mm}
Reinforcement Learning (RL) is commonly formalised as a Markov Decision Process (MDP)~\citep{sutton2018reinforcement} represented by the tuple $(\mathcal{S}, \mathcal{A}, \mathcal{P}, \mathcal{R})$, where $\mathcal{S}$ and $\mathcal{A}$ are the set of states and actions.
$\mathcal{P}(s_{t+1}|s_t, a_t)$ is the probability of transition from state $s_t$ to $s_{t+1}$ given an action $a_{t}$, where $s_t$, $s_{t+1} \in \mathcal{S}$  and $a_t \in \mathcal{A}$.
The reward function defines the reward obtained at each timestep $r_t=r(s_t, a_t, s_{t+1})$ when transitioning from state $s_t$ to $s_{t+1}$ under action $a_{t}$. 
An agent acting in the environment selects its next action based on the current state $s_t$ by following a policy $\pi_{\theta}(a_{t} | s_t)$.
The conventional objective in RL is then to optimize the parameters $\theta$ of policy $\pi_{\theta}$, such that it maximizes the expected cumulative reward $R(\tau)=\sum_{t=1}^T{r_t}$ over the entire episode trajectory $\tau$:
\begin{align}
    J(\pi_\theta) = \mathbb{E}_{\tau\sim\pi_\theta}\left[{\mathcal{R}(\tau)}\right]
\end{align}

% Policy Gradient objective?

% Model based
The transition probabilities $\mathcal{P}(s_{t+1}|s_t, a_t)$ of the environment are usually assumed to be unknown. Model-based RL~\citep{wang2019benchmarking} methods learn a parametric model $p_\phi(s_{t+1}|s_t, a_t)$ typically using supervised learning, from data collected when interacting in the environment.
Policies are then trained using transitions obtained by rolling out the model.

\newcommand\mycommfont[1]{\small\ttfamily\textcolor{blue}{#1}}
\SetCommentSty{mycommfont}
\begin{algorithm}[t]
\caption{\namedaqd{} and \name{} (\hlgray{highlighted lines} are specific to \name{})} 
\label{psuedocode}
    \textbf{Inputs:} $\numiterations{}$ num iterations, $\numimagiterations{}$ num imagined iterations, \hlgray{$p_{gradient}$ prop. of gradient-updated policies, and $p_{model} = 1-p_{gradient}$} (for \namedaqd{} $p_{gradient} = 0$ and $p_{model} = 1$)
    
    \textbf{Initialisation:} $\pop{}_0 \gets \text{init\_population()}, q_\phi \gets \text{init\_dynamics\_model()} ,$ \hlgray{$Q_\psi \gets \text{init\_critic()}$}
    
    \For{$\iter{} = 1, ..., \numiterations{}$}{
    
        $\imagpopit{} \gets \popit{}$ \tcp{copy $\popit{}$ in $\imagpopit{}$}
        
        \tcp{Optimize population in imagination}
        \For{$it{}_{imagination} = 1, ..., \numimagiterations{}$}{
            $\theta \gets \text{random\_selection}(\imagpopit{})$
            
            $\tilde{\theta} \gets \text{perturb}(\theta)$
            
            $F(\tilde{\theta}), d(\tilde{\theta}) \gets \text{evaluate\_imagination}(\tilde{\theta}, q_\phi)$
            
            $\imagpopit{} \gets \text{update\_population}(\tilde{\theta}, F(\tilde{\theta}), d(\tilde{\theta}))$
            }
        
       $\theta_{new} \gets \text{get\_last\_added}( \imagpopit{})$ \tcp{get last policies added to $\imagpopit{}$}
        
        $\theta_{model} \gets \text{select}(\theta_{new}, p_{model})$
        
        \hlgray{$\theta_{gradient} \gets \text{apply\_gradient}(Q_\psi, \text{select}(\theta_{new}, p_{gradient}))$}
        
        $\theta_{final} \gets (\theta_{model}, \theta_{gradient})$ \tcp{concatenate $\theta_{model}$ and $\theta_{gradient}$}
        
        %\tcp{Evaluate in environment, update all elements}
        $F(\theta), d(\theta) \gets \text{evaluate}(\theta_{final})$ \tcp{ evaluate to get reward $F$ and descriptor $d$}
        
        $\pop{}_{\iter{}+1} \gets \text{update\_population}(\theta, F(\theta), d(\theta))$
        
        $q_\phi \gets \text{update\_dynamics\_model}(q_\phi)$, \hlgray{$Q_\psi \gets \text{update\_critic}(Q_\psi)$}
        
        }
\Return $\pop{}_{\numiterations{}}$
\end{algorithm}

\subsection{Quality Diversity} \label{sec:qd}
\vspace{-2mm}
Quality-Diversity (QD) \citep{pugh2016quality, cully2017quality} are diversity-seeking population-based approaches to learning.
% QD methods improve a population $\pop{}$ to maximize the performance of each policy $\theta \in \pop{}$, and its diversity with respect to one another.
QD methods maintain a diversity of policies in the population $\pop{}$ while maximizing the performance of each policy $\theta \in \pop{}$. The population usually contains thousands of policies. 
QD considers an objective function $F(\theta)$ acting on the parameters of the policy $\theta$. 
Additionally, QD also considers a behavior descriptor $d(\theta)$ that characterizes the behavior induced by a policy. $d(\theta)$ is used to maintain solutions in their behavioral niche to guarantee the population diversity and that there is no two solutions in the population with similar behavior descriptor $d(\theta)$.
When applying QD to a RL problem, they are defined as follows, where $d(\tau)$ is the behavior descriptor of a given trajectory $\tau$:
\begin{align}
    F(\theta) = J(\pi_\theta) = \mathbb{E}_{\tau\sim\pi_\theta}\left[{\mathcal{R}(\tau)}\right] 
    \quad \text{and} \quad
    d(\theta) = \mathbb{E}_{\tau\sim\pi_\theta} \left[ d(\tau) \right]
\end{align}

Similar to Evolutionary Strategies (ES) \citep{salimans2017evolution}, QD methods operate on entire episodes and hence both the objective and the descriptor can be computed simultaneously for any parameter vector $\theta$.
QD methods then aim to maximize:
\begin{align}
   \max_{\pop{}} \quad \textrm{QD-Score}(\pop{}) =  \sum_{\theta \in \pop{}}{F(\theta)}
\end{align}
By maintaining a population $\pop{}$ of both diverse and high-performing policies, QD uses the existing parameters in the population as stepping stones~\citep{nguyen2016understanding, wang2019poet} in the optimization.
At each iteration $t$, a random set of policies are sampled from the current population $\popit{}$, perturbed and evaluated in the environment. 
Based on the results of these evaluations, these perturbed policies might replace existing ones or fill in a new niche in the population.
This incrementally improves the $\textrm{QD-Score}(\popit{})$ and encourages creativity and exploration.

\section{Method}
\label{sec:method}
\vspace{-2mm}
In this section, we introduce our two new model-based QD methods: \namedaqd{} and \name{}.

% DAQD
\subsection{\namedaqd{}: Learning and Sieving in Imagination} \label{subsec:method_1}
\vspace{-2mm}
QD approaches are conventionally driven by random perturbations of the parameters of solutions.
This undirected and divergent process gives QD its exploration capabilities but its major drawback is the number of samples it requires. %during the optimization detailed in Section \ref{sec:qd}. 
% making it hardly comparable to gradient-based approaches.
This is especially evident for high-dimensional optimization problems such as optimizing deep neural networks where usually more directed gradient based methods are used.
In this work, inspired by \cite{lim2021dynamics}, we perform the perturbation process in imagination, relying on a learned dynamics model of the environment to reduce the number of environment interactions when evaluating such perturbed policies.

At each iteration $j$ of the algorithm, the current population $\popit{}$ is first "copied" into imagination $\imagpopit{}$.
The policies of this provisional population $\pi_{\tilde{\theta}} \in \imagpopit{}$ are then perturbed, and evaluated in imagination using the rollouts of the dynamics model $q_\phi$.
Both the objective $F(\pi_{\tilde{\theta}})$ and the descriptor $d(\pi_{\tilde{\theta}})$ of the policies can be obtained from the state information present in the rollouts.
Using this process, $\imagpopit{}$ undergoes multiple steps of QD optimization in imagination.
The resulting policies $\pi_{\tilde{\theta}} \in \imagpopit{}$ that are added to the provisional population $\imagpopit{}$ during learning in imagination are then evaluated in the environment and used to update the population $\popit{}$ if they improve the QD-Score of the population.
The updated population $\popit{}$ is then used as a start for the next iteration of the algorithm.
This process of performing QD in imagination acts as a sieve and filters out perturbed solutions that are not likely to improve the quality and diversity of the population, hence increasing the sample efficiency.

Following \cite{chua2018deep}, we use a probabilistic bootstrap ensemble of models $q_\phi$ to capture uncertainties. 
Each model in the ensemble is a probabilistic model which predicts parameters of a Gaussian distribution $N(\mu_\phi(s_t, a_t), \Sigma_\phi(s_t, a_t))$ which we can then sample from, capturing the aleatoric uncertainty.
This model-based QD method corresponds to DA-QD~\citep{lim2021dynamics}. However,
we optimize high-dimensional closed loop neural network policies in complex exploration domains and hence, refer to it as \namedaqd{}.
%The ensemble of models.

%process acts as a data generator, harvesting a large diversity of transitions, used to get a better estimate of the dynamic of the environment through the model.
% This approach allows maintaining the exploration capabilities of QD by performing exploration in imagination.
% IMPORTANT TO STRESS SOMEWHERE - NOT SURE WHERE (COULD BE HERE OR IN RELATED WORK) THAT ORIGINAL DA-QD DOES NOT CONSIDER NN CONTROLLERS, ONLY SMALL OPEN LOOP CONTROLLERS. HERE WE SCALE THIS UP TO DEEPRL/NEUROEVOLUTION. OR WE COULD TALK LIKE IT IS VERY NEW ALSO and just put this fact in the related work 

\begin{figure*}[t!]
\centering
    \includegraphics[width = 0.99\textwidth]{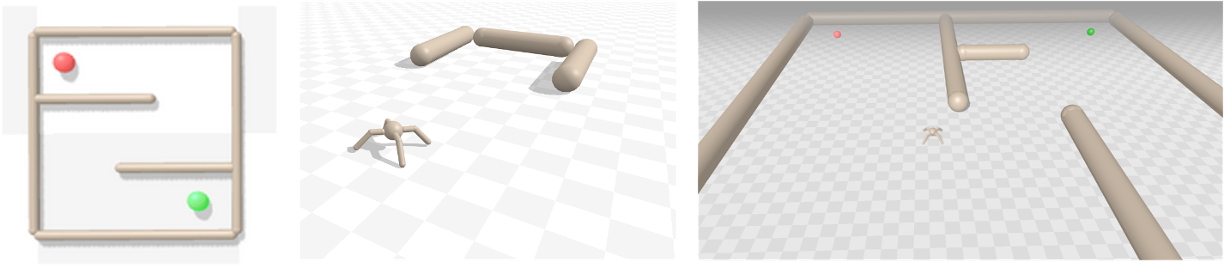}
    \caption{
        Deceptive reward tasks in the form of PointMaze, AntTrap and AntMaze environments.
    }
\label{fig:environments}
\end{figure*}

%PGA
\subsection{\name{}: Incorporating gradients in Quality-Diversity} \label{subsec:method_2}
\vspace{-2mm}
As mentioned above, the random perturbations driving QD approaches prove inefficient when applied to high-dimensional search spaces such as the parameters of deep neural networks~\citep{colas2020scaling}. 
To deal with this, we augment the usual perturbation operator with policy gradient information as done by~\citet{nilsson2021policy}.
The policy gradient can be more intuitively thought of as a more directed perturbation, hence being a more efficient optimization update procedure.
To apply policy gradients to a populationx, we maintain a critic network $Q_\psi$ which approximates the action-value function $Q(s_t, a_t)=\mathbb{E}\left[\sum_{k=0}^{T-t} \gamma^{k} r_{t+k+1} \mid s_t, a_t~\right]$ and gives the expected return from being in state $s_t$ and following action $a_t$. 
The critic allows us to gradually improve any policy in the direction maximizing the expected return by computing policy-gradient in Equation~\ref{eqn:policy_gradient}, approximated over a batch of transitions.
We train $Q_\psi$ using the same procedure as TD3~\citep{fujimoto2018addressing}.

\begin{equation} \label{eqn:policy_gradient}
    \nabla_{\theta_i} J(\theta_i)=\mathbb{E}_{\mathbf{s}, \mathbf{a} \sim \pi_{\theta_i}}\left[ \nabla_{\theta_i} \pi_{\theta_i}(\mathbf{s}) \left. \nabla_{\mathbf{a}} Q_{\psi}(\mathbf{s}, \mathbf{a}) \right. \right]
\end{equation}

\begin{figure*}[t!]
\centering
    \includegraphics[width = \textwidth]{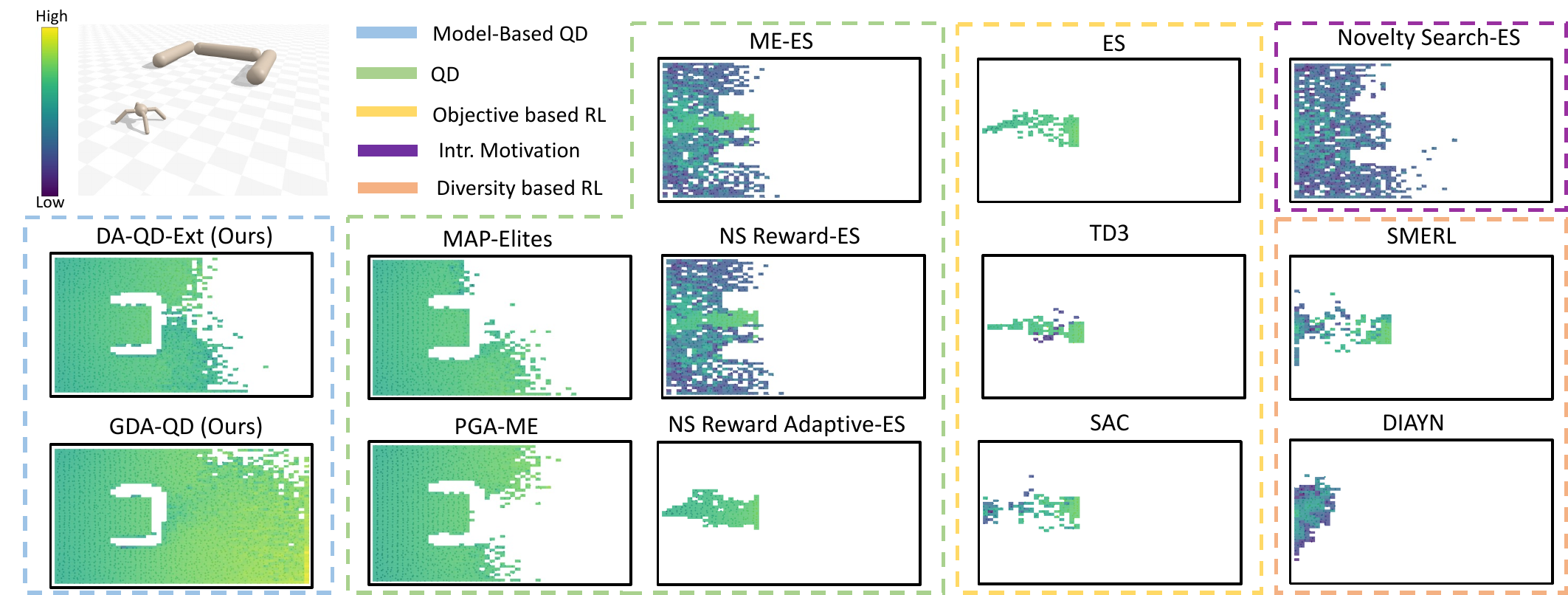}
    \caption{
        Final population of methods in the AntTrap environment (top left). 
        Each policy in the population is represented with a dot in the final position it manages to reach within the episode.
        The color of the square around each policy indicates its total reward, the lighter the better.
        The obstacle clearly appears on this plot, as the empty area in the middle.
        Our approach \name{} does not get stuck in the trap and clearly outperforms baselines given the same number of evaluations.
    }
\label{fig:archives}
\vspace{-4mm}
\end{figure*}

In \namedaqd{} explained in Section~\ref{subsec:method_1}, the population of policies $\popit{}$ is copied in imagination $\imagpopit{}$ to undergo multiple steps of QD optimization and returns $\batchsize{}$ policies to be evaluated.
In \name{}, we combine the efficient parameter based perturbation in imagination from~\namedaqd{} with more directed policy gradient updates explained above. The former is critical for efficient exploration, while the latter is critical for efficient exploitation.
To achieve this, a proportion $p_{model}$ of $\batchsize{}$ is sampled to be evaluated directly while another $p_{gradient} = 1 - p_{model}$ is sampled to be improved using the policy gradient operators before being evaluated.
% In imagination, only random perturbations are used to encourage exploration. % and optimize the population $\imagpopit{}$. 
In the following, we use $p_{model} = 0.9$.
We study the impact of this value and illustrate the complementary properties of these two generation-procedures in Appendix \ref{sec:ablation_pg_out}. 
We demonstrate that both parameter update procedures are essential to guarantee performance.
%We refer to this new algorithm as \name{} as it leverages both gradient information and learns the dynamics of the environment.
Figure \ref{fig:main_figure} and Algorithm \ref{psuedocode} provides a summary of this algorithm.

A key design choice in~\name{} for simplicity is that only parameter-based perturbations are applied in imagination, while the policy gradient updates are not applied in imagination. This was to clearly make a separation between efficient exploration and efficient exploitation (through gradient optimization).
Applying policy gradient updates in imagination could potentially further improve \name{} at the cost of training the critic more often in imagination. We leave this for future work.
% At each iteration, \name{} uses the dynamical model to generate a fixed number of candidate policies in imagination. 
% Among these policies, $p_{model}$ are policies that prove promising while evaluated in imagination, and $1 - p_{model}$ are generated by improving such solutions using policy-gradient.
%In the following, we perturb $10\%$ of the total number of policies returned at each iteration using policy-gradient, which corresponds to $p_{model} = 90\%$. 

% ES/Others
% In tasks where exploration is important, usually longer episodes with more time steps.
% ES is more effective than policy gradient methods when dealing with more time steps~\citep{salimans2017evolution}
% large sample sizes to be effective and a competitive alternative to policy gradient approaches and compute resources (Number of CPUs)
% model based also alleviate this.

\subsection{Quality-Diversity as data generators for deep models}
\vspace{-2mm}
A key quality of QD algorithms is that they are excellent at generating diverse and high-quality data~\citep{ecoffet2021first,grillotti2022unsupervised}.
In our work, the search for a diversity of high-performing behaviors when optimizing the population results in a diverse dataset of transitions.
We leverage this property to train deep models which require and often excel when provided with such data.
As commonly done, the diverse dataset of transitions is stored in a replay buffer and used to train (1) the dynamics model $q_\phi$ and (2) the critic $Q_\psi$.
Both these models are suitable candidates as the training of the critic $Q_\psi$ is off-policy and can trained using transitions collected from any behavioral policy. 
Additionally, training the dynamics model $q_\phi$ is a supervised learning problem  which would benefit from a large and diverse dataset of transitions.

We found this property to be especially important to our method.
Specific to the training of the dynamics model, \name{} does not use transitions produce by gradient-optimized policies to train the model.
This was found empirically in our studies as we observed the transitions produced by the gradient-optimized policies induced a shift in distribution that prove detrimental for the training of the dynamics models.
Results of this study are detailed later in Section \ref{sec:filter_transitions}. 
It is important to recognize that the transitions used to train the dynamics model are not merely just from policies that have been randomly perturbed but a population of policies that have undergone multiple steps of QD optimization in imagination and are expected to improve the population. This results in a diverse and high performing dataset of transitions that \name{} uses to further train its models.

\section{Experiments}
\label{sec:exp}
\vspace{-2mm}
We aim to evaluate our method by answering four main questions: 
(1) Can we scale model-based QD approaches to RL domains and Neuroevolution? 
(2) Does \name{} results in more performant and sample-efficient learning than traditional QD approaches and simple model-based QD?
(3) What is the importance of the policy-gradient perturbation in the performance of \name{}?
(4) How does the data-generation capabilities of \name{} enforce efficient learning?

% Can we scale DA-QD to deep nueroevolution?
% Does \name result in more performant and sample effieinct learning?

% \begin{table}[]
% \tiny
% \centering
% \begin{tabular}{l l | c c c c c c c c| c c | c c c c}
% \toprule
%  & &
%  & \textsc{\namedaqd{}} 
%  & \textsc{\name{}} 
%  & \textsc{ME} 
%  & \textsc{PGA-ME}
%  & \textsc{ME-ES}
%  & \textsc{NSR-ES} 
%  & \textsc{NSRA-ES} 
%  & \textsc{ES} 
%  & \textsc{NS-ES} 
%  & \textsc{TD3}
%  & \textsc{SAC}
%  & \textsc{DAYN}
%  & \textsc{SMERL} 
%  \\ 
% \midrule
%  \textsc{AntTrap} & \textsc{QD-Score} & & & & & & & & & & & & & \\

% \end{tabular}
% \caption{
%         Final QD-Score, Coverage and Max-Total-Reward reached by all algorithms on the Ant-Trap task for $10^6$ policies evaluations.
%         Each experiment is replicated $3$ times, we report in the table the median value across runs.
% }
% \label{tab:baselines}
% \end{table}

\begin{table}[]
\scriptsize
\centering
\begin{tabular}{l|ccc|ccc|ccc}
\toprule
 & \multicolumn{3}{c}{\textsc{PointMaze}}
 & \multicolumn{3}{c}{\textsc{AntTrap}}
 & \multicolumn{3}{c}{\textsc{AntMaze}}
 \\ 
 & \textsc{QD-Score}
 & \textsc{Cov}
 & \textsc{Max-Rew} 
  & \textsc{QD-Score}
 & \textsc{Cov}
 & \textsc{Max-Rew} 
  & \textsc{QD-Score}
 & \textsc{Cov}
 & \textsc{Max-Rew} 
 \\ 
\midrule
\addlinespace
TD3 
& - & - & -126.38
& - & - & 189.52
& - & - & 1.05            \\ 
SAC 
& - & - & -126.18   
& - & - & 204.68 
& - & - & 1.06         \\ 
ES 
&  0.46 &  0.52 &  -126.85                
& 2.97 & 2.91 & 200.95
& 18.6 & 10.88  & 0.97
\\ 
\midrule
\addlinespace
DIAYN 
& - & - & -67.98                
& - & - & -6.15 
& - & - & 0.20        \\ 
SMERL 
&  -  &  - &  -38.29
&  -  &  - &  171.81
&  -  &  - &  1.06     \\ 
\midrule
\addlinespace
NS-ES 
& 0.93 & 1.8 & -147.80                  
& 10.12 & 28.76 & -13.14
& 45.45 & 42.6 & 1.23
\\ 
\midrule
\addlinespace
ME 
& 93.74 & \textbf{99.92} & -25.36         
& 42.44 & 43.44 & 218.77
& 56.71 & 37.68 & 1.29
\\ 
PGA-ME           
&  93.06 &  \textbf{99.92} &  \textbf{-24.06}      
& 47.82 & 47.08 & 274.52
& 62.56 & 39.66 &  1.48
\\
NSR-ES         
& 1.07 & 1.24 & -126.85                  
& 6.36 & 6.26 & 196.30
& 22.89 & 14.24 & 1.02
\\
NSRA-ES         
& 1.44 & 1.78 & -126.85                 
& 14.18 & 29.32 & 170.38
& 46.5 & 43.56 & 1.30
\\
ME-ES         
& 21.0 & 30.52 & -62.05   
& 12.70 & 21.92 & 157.42
& 38.5 & 33.48 & 1.15
\\
\namedaqd{}    
& 96.67 & \textbf{99.92} & \textbf{-24.81}
& 50.33 & 51.0 & 196.66 
& 70.94 & 43.7 & 1.51
\\ 
\name{}   
& \textbf{97.70} & \textbf{99.92} & \textbf{-24.24}
& \textbf{76.28} & \textbf{72.44} & \textbf{342.24}  
& \textbf{80.5} & \textbf{51.4} & \textbf{1.87}
\\
\bottomrule
\end{tabular}
\caption{
        Final QD-Score ($\%$ of maximum value), Coverage ($\%$) and Max-Total-Reward reached by all algorithms on all considered tasks.
        Each experiment is replicated $15$ times, we report in the table the median value across runs. In the algorithms name, ME stands for MAP-Elites.
}
\label{tab:baselines}
\vspace{-3mm}
\end{table}

\subsection{Experimental setup}
\vspace{-2mm}
% Tasks and Environments
\textbf{Tasks and Environments:}
% We focus on the AntTrap domain~\citep{conti2018improving, colas2020scaling, cideron2020qd}, considered in the literature as an hard exploration problems. In this task, an 8-DoF Ant robot learns how to walk, aiming to go beyond the trap~(see Figure~\ref{fig:archives}). It gets increasing rewards for going as fast as possible in a straight line while minimizing energy-usage. This reward definition makes this task deceptive as it drives the robot directly into the trap.
% The descriptor $d(\pi_{\theta})$ is defined as the x-y position at the end of the trajectory.
We focus on tasks considered in literature as hard exploration problems: PointMaze~\cite{lehman2011novelty, parker2020effective}, AntTrap~\citep{conti2018improving, colas2020scaling, cideron2020qd, parker2020effective} and AntMaze~\citep{colas2020scaling, cideron2020qd, salter2022mo2} (see Fig. \ref{fig:environments}). 
The reward in these tasks is deceptive making exploration and diversity critical when solving them.
To start, we consider a simple PointMaze environment where a 2-dimensional point agent is given a reward corresponding to the distance to the goal in the maze.
The AntTrap and AntMaze are higher dimensional continuous control tasks where an 8-DoF Ant robot learns how to walk, aiming to go beyond the trap in AntTrap and to reach the goal in AntMaze. In AntTrap, the robot gets increasing rewards for going as fast as possible while minimizing energy-usage. In the AntMaze tasks, the reward is the distance to the goal.
This reward definition for all the tasks considered makes them deceptive.
The descriptor $d(\pi_{\theta})$ used in all the tasks is defined as the x-y position at the end of the trajectory $(x_T, y_T)$.

% Baselines
\textbf{Baselines:} Across our experiments, we consider the following baselines:\vspace{-0.5em}
\begin{itemize}[leftmargin=*] \setlength\itemsep{0em}
    \item \textbf{MAP-Elites:} the most-commonly used QD algorithm~\citep{mouret2015illuminating}. 
    \item \textbf{PGA-MAP-Elites:}~\citep{nilsson2021policy}, augments MAP-Elites with a policy-gradient based update operator.
    \item \textbf{OpenAI ES:} Evolution Strategy~\citep{salimans2017evolution} relying on natural gradient approximation.
    \item \textbf{Novelty Search ES:} We compare against NS-ES~\cite{conti2018improving} as an intrinsic motivation baseline. This uses the OpenAI-ES algorithm but with a novelty reward instead of task reward..
    \item \textbf{QD-ES Algorithms:} QD-ES algorithms consider both quality and novelty during optimization. We use NSR-ES, NSRA-ES~\citep{conti2018improving}, and MAP-Elites-ES (ME-ES)~\citep{colas2020scaling}. NSR-ES and NSRA-ES build on NS-ES by including the task reward term as a weighted sum with the novelty reward term.
    ME-ES mixes MAP-Elites with OpenAI ES ~\citep{colas2020scaling}.
    \item \textbf{Single Policy Deep RL Algorithms:} We consider TD3~\cite{fujimoto2018addressing} and SAC~\cite{haarnoja2018soft}. SAC is entropy regularized and is a popular choice for greater exploration.
    \item \textbf{Mutual Information RL Algorithms:} We also consider DIAYN~\cite{eysenbach2018diversity} and SMERL~\cite{kumar2020one} which are also diversity seeking algorithms. To ensure the comparisons are fair with descriptor based methods, we use the x-y prior when running these algorithms. 
    DIAYN is purely unsupervised and does not consider the task rewards. SMERL considers the task reward during optimization
\end{itemize}
For fairness, as the single policy baselines (TD3, SAC, DIAYN, SMERL) do not rely on a population, nor on complete-episode evaluations, we only reports its final value in the results as a dotted line. 
For a more qualitative comparison of these algorithms, we also collect the trajectories of the policies throughout the learning process and plot them as part of a population. To ensure the comparability of algorithms using ES (OpenAI-ES, NS-ES, NSR-ES, NSRA-ES, ME-ES) with other approaches, we consider every estimate-evaluation as one sample, making these algorithms highly sample-inefficient.

% Metrics
\textbf{Metrics:} We consider two metrics to assess the performance of \name{}: \vspace{-0.5em}
\begin{itemize}[leftmargin=*]\setlength\itemsep{0em}
    \item \textbf{QD-Score:} defined in Section \ref{sec:qd}. It quantifies the diversity and quality of the overall population and allows to compare population-based methods.%population-wide metrics.
    % \item \textbf{Coverage:} Coverage is defined as a percentage of the descriptor space filled by the population, used as a measure of the diversity of the population.
    \item \textbf{Max-Total-Reward:} the total reward of the best individual of the current population. This metrics allows comparison with single-policy methods such as RL baselines.
\end{itemize}

\begin{figure*}[t!]
\centering
    \includegraphics[width = 0.99\textwidth]{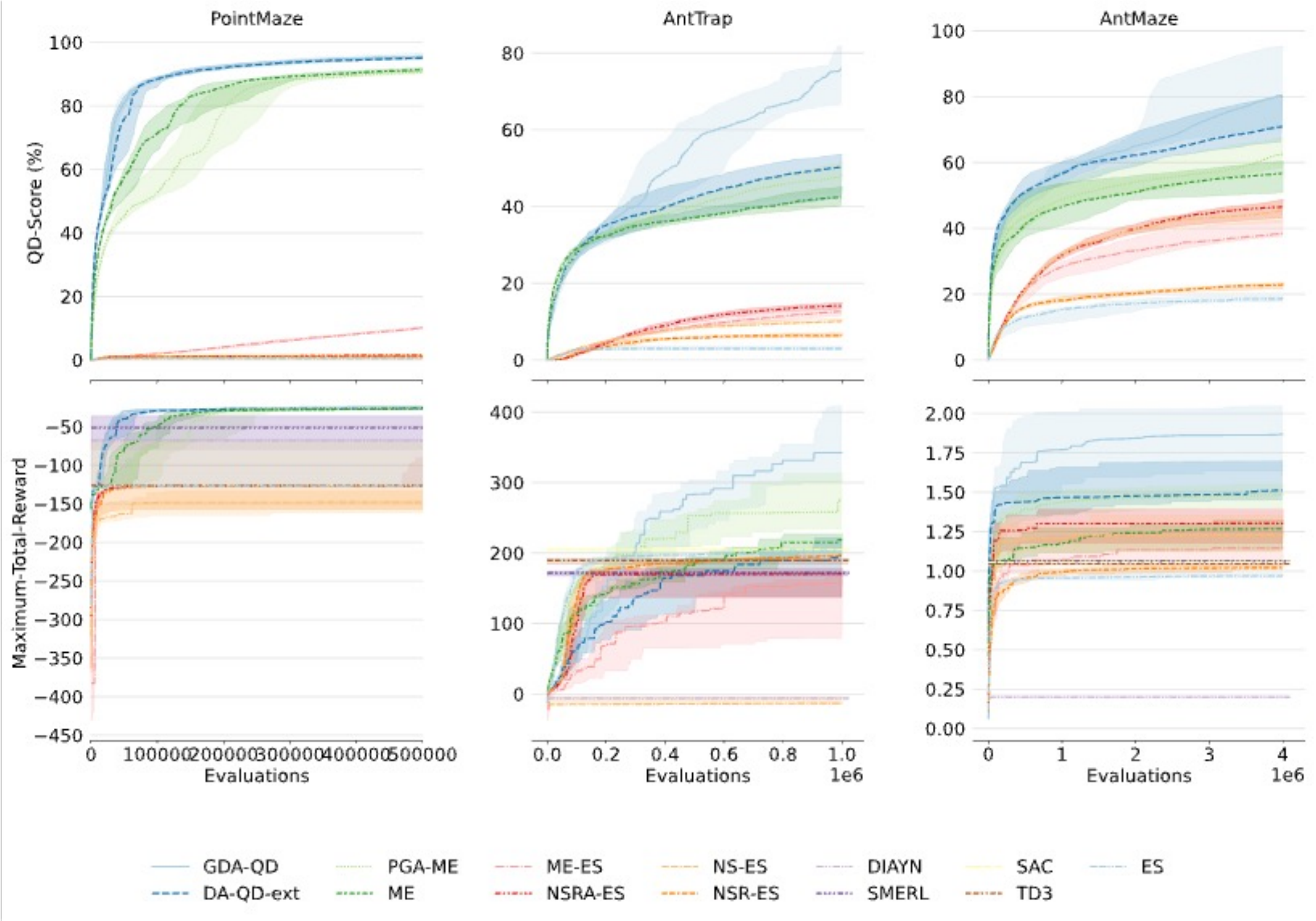}
    \caption{
        QD-Score (top) Max-Total-Reward (bottom) of all algorithms on the AntTrap (left), PointMaze (centre) and AntMaze (right) tasks plotted against number of evaluations. 
        Each experiment is replicated $15$ times, the solid line corresponds to the median over replications and the shaded area to the first and third quartiles.
    }
\label{fig:graphs}
\end{figure*}

% Hyperparameters
\textbf{Implementation and hyperparameters:}
% Our source code is available at \sourcecode{}. \footnote{The code will be made available after the review period}.
% It includes a containerised environment to replicate our experiments.
The methods presented in this paper as well as all our baselines are based on the implementation of MAP-Elites in the QDax open-source library \citep{lim2022accelerated}, using the Brax simulator \cite{brax2021github}.
All hyperparameters and implementation details used in our algorithms and for model training can be found in the Appendix~\ref{app:hyperparams}.
% We use a batch-size of $512$ for MAP-Elites with a maximal population size of $2500$ policies.
% All the hyperparameters used to train the critic networks are taken from the original TD3 paper \cite{fujimoto2018addressing}. 
% For the dynamics model we use an ensemble of $4$ probabilistic models. The detailed hyperparameters can be found in Appendix \ref{app:hyperparams}.

\subsection{Results}
\vspace{-2mm}
The results of our experiments are summarized in Figure \ref{fig:graphs} and Table \ref{tab:baselines}.
We also display a visualization of the final populations of policies for each algorithm in Figure \ref{fig:archives} (AntTrap) and Appendix \ref{app:supp} (PointMaze and AntMaze).

Figure~\ref{fig:graphs} shows that both our proposed model-based versions, \namedaqd{} and \name{},  significantly outperform all baselines in terms of sample efficiency and final performance.
This demonstrates that we can scale model-based QD methods to deep neuroevolution.
The performance of \namedaqd{} suggests that learning diverse behaviors in imagination using the dynamics model is a simple but effective approach to save samples yet maintain the divergent search capabilities of the random perturbations.
However, we can see that the maximum total reward obtained by \namedaqd{} seems to stagnate and increase slowly, especially when compared to \name{} and PGA-MAP-Elites.
This can be explained by the absence of reward maximizing optimization updates such as the policy gradient updates present in \name{} and PGA-MAP-Elites.
Hence, \name{} is shown to get the "best of both worlds" by benefitting from the QD in imagination as well as the policy gradient updates. This is evident in its performance across the QD-Score and max. return metrics.

In terms of baselines, Figure\ref{fig:archives} show that all objective-only baselines: SAC, TD3 and ES, struggle with deceptive rewards. They all get stuck into the traps present in AntTrap and AntMaze. 
Diversity Based RL algorihtms such as DIAYN struggle with no task reward signal while SMERL manages to reach the final goal only in the simpler PointMaze.
The single-agent baselines (TD3, SAC, DIAYN, and SMERL) are plotted as horizontal lines to represent the max performance obtained by the agent. This is done to allow comparison as they do not rely on a population, making the Coverage and QD-Score metrics not relevant, and also performs policies-updates within episodes.
The poor performance of ES-based algorithms, in particular ME-ES, can be attributed to the number of samples ($\sim$ hundreds or thousands) required just to approximate a single gradient step.
It is important to note that these methods commonly do not consider the notion of evaluations and generally evaluate the algorithms versus time or number of generations as they are suited to be heavily parallelized across clusters of CPUs. 
Despite this limitation, intrinsically-motivated ES baselines NS-ES and its QD variant NSRA-ES manage to get good coverage on AntTrap and AntMaze. However, they struggle to discover high-performing solutions within the given evaluation budget. We provide a visualization of the adaptive mechanism of NSRA-ES in Appendix \ref{app:supp}.

\begin{figure*}[t!]
\centering
    \includegraphics[width = 0.99\textwidth]{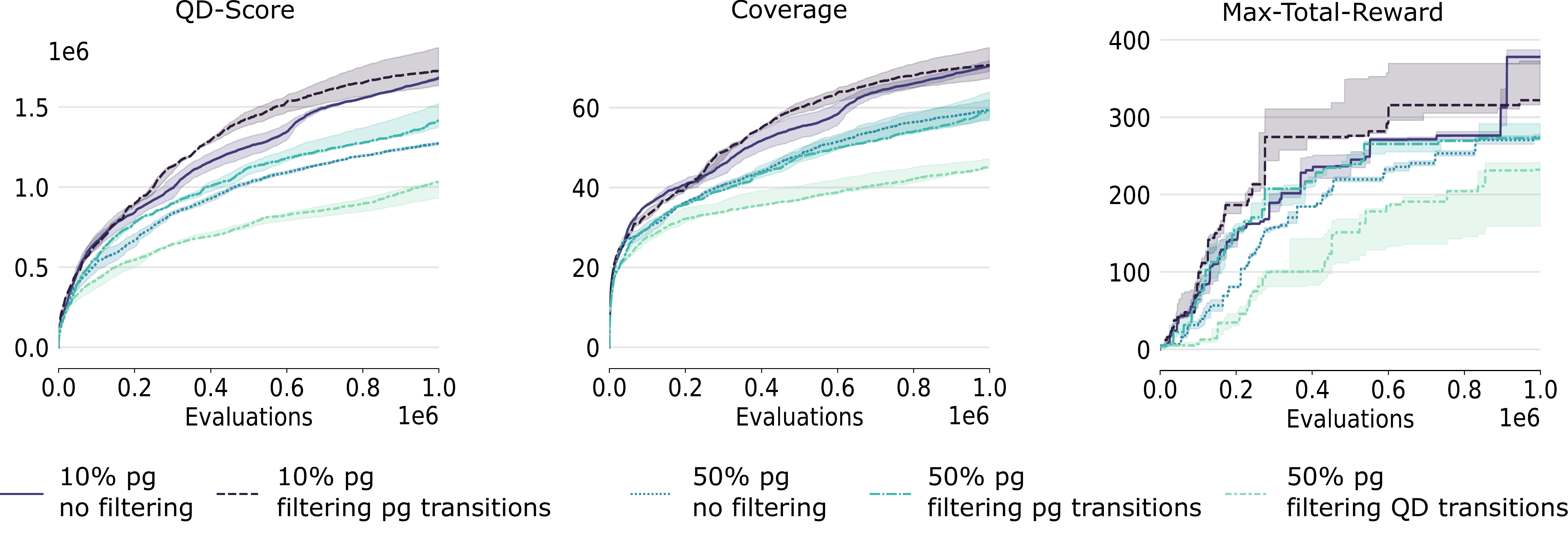}
    \caption{
        QD-Score (left), Coverage (middle) and Max-Total-Reward (right) on the AntTrap task of \name{} with different proportion of policy-gradients generated policies (pg) and different transitions filtering schemes.
        The solid line corresponds to the median over replications and the shaded area to the first and third quartiles.
    }
\label{fig:ablation_model}
\vspace{-3mm}
\end{figure*}

\subsection{Importance of QD for training deep models} \label{sec:filter_transitions}
\vspace{-2mm}
To investigate the data generation abilities of QD, we compare the performance of the algorithm when the dynamics model is trained on different data distributions based on the policies rolled out in the environment.
In our case, we have two main types of data generators: policies obtained thought the QD process in imagination (giving $D_{model}$), and policies perturbed also using policy-gradients (giving $D_{gradient}$).
We run an ablation where we train the dynamics model on transitions collected by either $D_{model}$, $D_{gradient}$, or a mixture of both $D_{model+gradient}$, by filtering out the corresponding transitions.
Figure~\ref{fig:ablation_model} shows the performance curves when running this ablation on the AntTrap task.
We first test the effect of training the model with $D_{model}$ when $p_{gradient}=0.1$. 
We notice a minor difference in which experiments that train only on $D_{model}$ perform better than when not using any filter (i.e. $D_{model+gradient}$).
To enable a fair comparison in terms of the number of transitions added to the replay buffer when attempting to do the converse (i.e. using $D_{gradient}$), we use a $p_{model} = p_{gradient}=0.5$.
We observe better performance when the dynamics model is only trained with $D_{model}$.
This is compared to training on $D_{model+gradient}$ where the difference is minimal as seen previously.
However, performance significantly drops when training just on $D_{gradient}$.
We hypothesize that this is due to bias in the transitions obtained through the policies perturbed by policy gradient resulting in a skewed dataset while the transitions given by policies obtained from QD in imagination provides a diverse and high-quality dataset.

\section{Related Work}
\label{sec:relatedwork}
\vspace{-2mm}
\textbf{Searching for Diversity.} 
Prior studies have shown the importance of maintaining a diverse set of solutions to solve a problem. 
Novelty-search approaches~\citep{lehman2011evolving} inspired from evolutionary computation optimize for the novelty of solutions defined by a behavioral characterization with respect to a population instead of the optimization objective.
QD methods~\citep{cully2017quality, mouret2015illuminating} extend this approach by also considering the objective, aiming to find both diverse and high-performing population of solutions.
Similarly, our work builds on QD approaches with the aim to maintain a diversity of high-performing solutions.
%Our work rely on MAP-Elites but \cite{lehman2011evolving} have proposed an alternative QD approach relying on a novelty archive.
Other approaches to searching for diversity also exist in the RL community.
Unsupervised RL methods~\citep{eysenbach2018diversity, sharma2020emergent} commonly use a mutual information maximization objective to learn a diversity of behaviors in a skill-conditioned policies.
Similar to QD,~\citet{kumar2020one, zahavy2022discovering} have proposed to extend these unsupervised RL approaches by integrating objectives using constrained Markov decision processes. 
% \cite{parker2020effective} define a metric to quantify the novelty of a policy based on the actions it generates.
%In our work, diversity is defined based on the states encountered by the agent but not on the full state distribution nor on the action distribution. 
While these approaches typically focus on maintaining a dozen of different policies, our algorithm discovers thousands of independent diverse policies.
%Similarly~\cite{zahavy2022discovering, kumar2020one} propose to use successor features to maintain a population of diverse sub-optimal policies and \cite{parker2020effective} define a metric to quantify the novelty of a policy based on the actions it generates.

\textbf{Neuroevolution} seeks to evolve neural networks through biologically-inspired methods such as evolutionary algorithms and have interesting properties unavailable to common gradient-based methods~\citep{stanley2019designing}.
However, a limitation in the neuroevolution domain is the dimensionality of the search-space, that quickly limits the effectiveness of random perturbations. 
Some methods have overcome this limitation by using indirect encoding methods~\citep{clune2009evolving} or through natural approximated gradients~\citep{salimans2017evolution}.
Our work hybridizes neuroevolution with deep reinforcement learning methods. Similar to~\citet{nilsson2021policy}, we use policy-gradients as directed perturbations to effectively maneuver the high-dimensional search space but, we significantly improve the sample efficiency and performance by also augmenting these operators with model-based methods.
% Neuroevolution has also been used in QD algorithms~\citep{colas2020scaling, nilsson2021policy, pierrot2021diversity}.
% While there has been work done to scale these methods to be used
% y.either augmented with gradients \citep{cideron2020qd, nilsson2021policy} or approximated gradients \citep{colas2020scaling}.

\textbf{Model-based Quality-Diversity.}
As the perturbations commonly used in QD are sample inefficient, prior work has sought the use of data-driven models to alleviate this.
SAIL~\cite{gaier2018data} introduced the use of a surrogate model in the form of a Gaussian Process model to predict the objective.
As Gaussian processes generally only work well on low-dimensional data, more recent methods have explored the use of deep networks~\citep{keller2020model, lim2021dynamics, zhang2022deep} as forms surrogate models to predict both the objective and descriptors.
We utilize the model-based QD framework from~\citet{lim2021dynamics} which first introduced the idea of maintaining an imagined population and also builds on model-based RL methods~\citep{wang2019benchmarking} where a dynamics model is used.
Critically, this work has only been applied to low-dimensional open-loop policies. 
To the best of our knowledge, our work is the first model-based QD algorithm that scales to the more complex deep neuroevolution domain where we optimize closed loop RL policies.

\section{Conclusion and Future Work}
\label{sec:conclusion}
\vspace{-2mm}
% Summary
In this paper, we introduce a novel model-based Quality-Diversity method, \name, which optimizes a population of diverse policies to explore more efficiently.
To the best of our knowledge, this approach is the first model-based QD algorithm scaling to neuroevolution to optimize deep neural network policies.
We leverage a key property of QD algorithms as effective data generators to train deep models in the form of a dynamics model and a critic.
In turn, these models help to significantly improve the sample efficiency and final performance of the QD algorithm.
The dynamics model is used to learn and sieve policies in imagination while the critic is used to apply policy gradient updates to sampled policies.
Our experiments show that \name{} outperforms a range of Deep RL and QD baselines on a hard exploration task containing deceptive rewards.
Overall, we demonstrate some of the powerful synergies that can arise between population-based learning and deep learning approaches.
% Limitations
In future work, we hope to extend our work to more complex domains through the use of latent dynamics models~\citep{ha2018recurrent, hafner2019dream}.

\begin{ack}
% Use unnumbered first level headings for the acknowledgments. All acknowledgments
% More information about this disclosure can be found at: \url{https://neurips.cc/Conferences/2022/PaperInformation/FundingDisclosure}.

% Do {\bf not} include this section in the anonymized submission, only in the final paper. You can use the \texttt{ack} environment provided in the style file to autmoatically hide this section in the anonymized submission.
This work was supported by the Engineering and Physical Sciences Research Council (EPSRC) grant EP/V006673/1 project REcoVER. 
\end{ack}

%\section*{References}
{
\small
\bibliographystyle{plainnat}
\bibliography{references}
}

\newpage
\appendix

% \section{Appendix}
\section{Implementation Details} \label{app:hyperparams}

\subsection{Algorithm Hyperparameters}

We give in Table \ref{tab:hyperparams} all the hyperparameters used for \name{}.

\begin{table}[htb]
\centering
\begin{tabular}{l|cccc}
\toprule
\textsc{Hyperparameter} & 
\textsc{MAP-Elites} & \textsc{PGA-ME} & \textsc{\namedaqd{}} & \textsc{\name{}} \\ 
\midrule
\addlinespace
Policy hidden layer sizes  & [64, 64] & [64, 64] & [64, 64] & [64, 64] \\
Batch size, $\batchsize{}$ & 512 & 512 & 512 & 512\\
\midrule
\addlinespace
Iso coefficient, $\sigma_1$  & 0.01 & 0.01 & 0.01 & 0.01 \\
Line coefficient, $\sigma_2$ & 0.1 & 0.1 & 0.1 & 0.1 \\
\midrule
\addlinespace
Max. imagined iterations, $\numimagiterations{}$& - & - & 100 & 100 \\
Size of add buffer $\addbuffer{}$ & - & - & 512 & 512\\
Surrogate hidden layer sizes & - & - & [512, 512] & [512, 512] \\
Surrogate ensemble size& - & - & 4 & 4 \\
Surrogate learning rate& - & - & 0.001 & 0.001  \\
Surrogate batch size& - & - & 512 & 512 \\
Max. model training steps, $\maxmodelsteps{}$ &  - & - & 2000 & 2000 \\
Surrogate replay buffer size& - & - & 4000000  & 4000000\\
Model update period, $\modeltrainperiod{}$& - & - & 25 & 25 \\
Max epochs since improvement& - & - & 10 & 10 \\
\midrule
\addlinespace
Proportion Mutation ga & - &  0.5 & - &  - \\ 
Num. critic training steps &  - & 300 & - & 300 \\ 
Num. PG training steps &  - &  100 &  - &  100  \\ 
PGA replay buffer size &  - &  1000000 & - &  1000000  \\
Critic hidden layer size &  - & [256, 256] &  - & [256, 256] \\
Critic learning rate & - &  0.0003 & - &  0.0003   \\ 
Greedy learning rate &  - & 0.0003  &  - & 0.0003      \\ 
PG update learning rate &  - & 0.001 &  - & 0.001 \\
Noise clip &  - & 0.5  &  - & 0.5 \\
Policy noise &  - & 0.2 &  - & 0.2 \\
Discount $\gamma$ &  - & 0.99 &  - & 0.99 \\
Reward scaling &  - & 1.0 &  - & 1.0 \\
Transitions batch size &  - & 256 &  - & 256 \\
Soft tau update &  - & 0.005 &  - & 0.005 \\
\midrule
\addlinespace
Proportion model, $p_{model}$ & - & - & - & 0.9 \\
\bottomrule
\end{tabular}
\caption{
    Hyperparameters of \name{} and baselines.
}
\label{tab:hyperparams}
\end{table}

\textbf{Perturbation Operators.} We use a directional variation by \citet{vassiliades2018iso} which is defined by the equation:
\begin{equation} \label{equation:directional-variation}    
\tilde{\theta} = \theta_1 + \sigma_1 \mathcal{N}(0, I) + \sigma_2 \mathcal{N}(0, 1) (\theta_{2} - \theta_{1})
\end{equation}

where $\theta_{1}$ and $\theta_{2}$, are the parameters of two policies from the population. Perturbed parameters, $\tilde{\theta}$, are obtained by adding Gaussian noise with a covariance matrix $\sigma_1 N(0, I)$ to $\theta_{1}$ controlled by isolated noise coefficient $\sigma_1$. The resulting vector is then moved along the line from $\theta_{1}$ to $\theta_{2}$ by line coefficient $\sigma_2$. These hyperparameters can be found in Table~\ref{tab:hyperparams} and have been found empirically.

\textbf{DA-QD details.}
QD is performed in imagination until the add buffer $\addbuffer{}$ is filled or until a maximum number of imagined iterations $\numimagiterations{}$. This is to prevent the algorithm being stuck in imagination especially when the algorithm is close to convergence.

The dynamics model is trained every $\modeltrainperiod{}$ iterations of the full QD loop.

\textbf{Dynamics model details.}
The hyperparamters of the model architecture can be found in Table~\ref{tab:baselines}. 
We train the model by minimizing the negative log-likelihood as done in~\citet{chua2018deep}.
Everytime the model is trained, the replay buffer is randomnly split into a train and validation sets. The model is trained until the the improvement of the validation loss is below $1\%$ or until a maximum number of gradient steps is reached $\maxmodelsteps{}$. The dynamics model rollout length in imagination is always equivalent to the corresponding episode length of the environment.

\textbf{Policy Gradient operator details}
We use the hyper-parameters from the PGA-ME~\cite{nilsson2021policy} for the training of the TD3 agents (both actor and critic components). Greedy here refers to the base actor policy/agent in TD3. PG learning rate here refers to the learning rate used when applying the policy gradient perturbations to policies sampled from the population.

\subsection{Environment Hyperparameters}

We provide the different hyper-parameters used for the different environments considered.

\begin{table}[htb]
\centering
\begin{tabular}{l|ccc}
\toprule
\textsc{Hyperparameter} & \textsc{PointMaze}  & \textsc{AntTrap}  & \textsc{AntMaze} \\ 
\midrule
\addlinespace
Episode Length, $T$ & 250 & 200 & 500 \\
Evaluation Budget, $\numiterations{}$ & 1,000,000 & 1,000,000 & 4,000,000\\
Population size & 2500 & 2500 & 2500 \\
\bottomrule
\end{tabular}
\caption{
    Environment hyperparameters used.
}
\label{tab:env_hyperparams}
\end{table}

\section{Ablation of the proportion of policy-gradient} \label{sec:ablation_pg_out}

\begin{figure*}[h]
\centering
    \includegraphics[width = 0.99\textwidth]{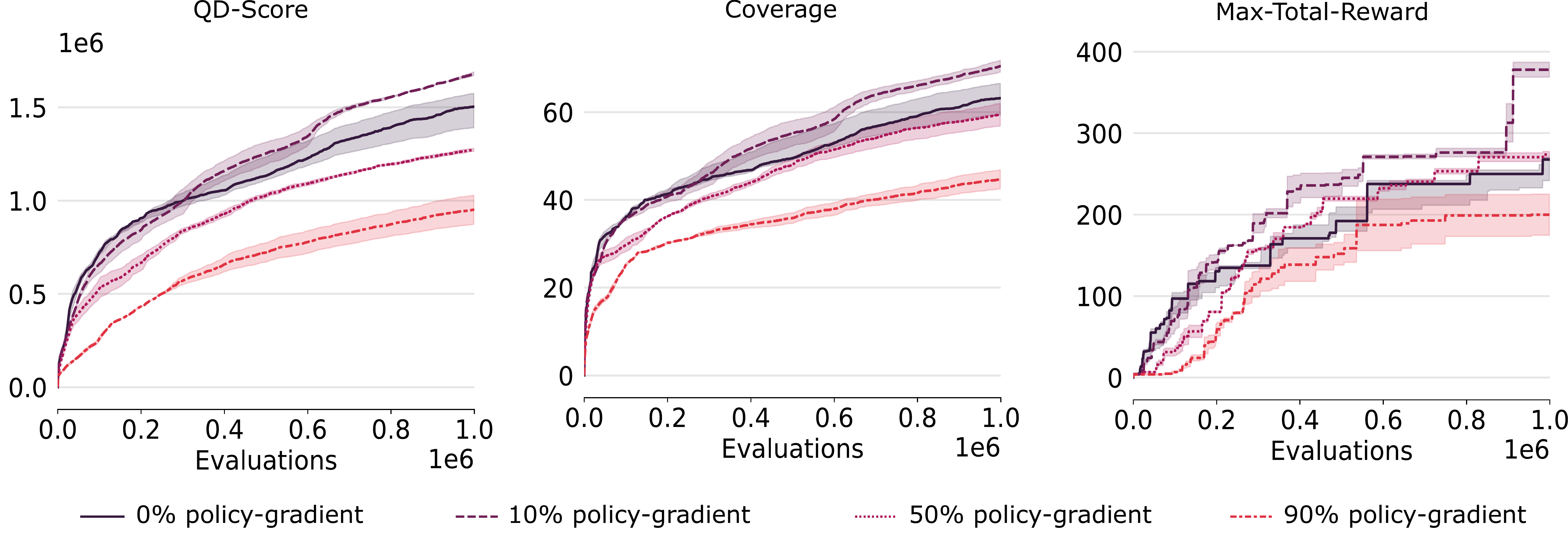}
    \caption{
        QD-Score (left), Coverage (middle) and Max-Total-Reward (right) on the AntTrap task of \name{} with different values of $p_{gradient}$. 
        $p_{gradient}$ gives the proportion of policies improved using policy-gradients.
        Each experiment is replicated $3$ times, the solid line corresponds to the median over replications and the shaded area to the first and third quartiles.
    }
\label{fig:ablation_pg}
\end{figure*}

In this section, we aim to demonstrate the importance of both types of perturbations used in \name{}: policy-gradient and random model-filtered perturbations.
We run an ablation of the different proportions $p_{model}$ and $p_{gradient}=1-p_{model}$ of perturbations at each generation.
In Figure~\ref{fig:ablation_pg}, we can see that the configuration of $p_{gradient}=0.1$ performs the best.
This indicates that while the policy gradient operators are important, they are also generally wasteful evaluations.
This is further backed up by the fact that the performance deteriorates as the proportion $p_{gradient}$ increases over $p_{model}$.
However, it is also still key to maintain some proportion of policy gradient updates $p_{gradient}$, as not using them is detrimental to the max total reward which results in an overall lower performance of the population $\pop{}$.

% Figure~\ref{} shows a plot which shows the proportion of each the different perturbation operators improves the population $\pop{}$ either by being more novel and better-performing than existing policies in the population.

\newpage 
\section{Supplemetary Results} \label{app:supp}
\vspace{-2mm}
\subsection{Visualization of population}
\begin{figure*}[b!]
\centering
    \includegraphics[width = \textwidth]{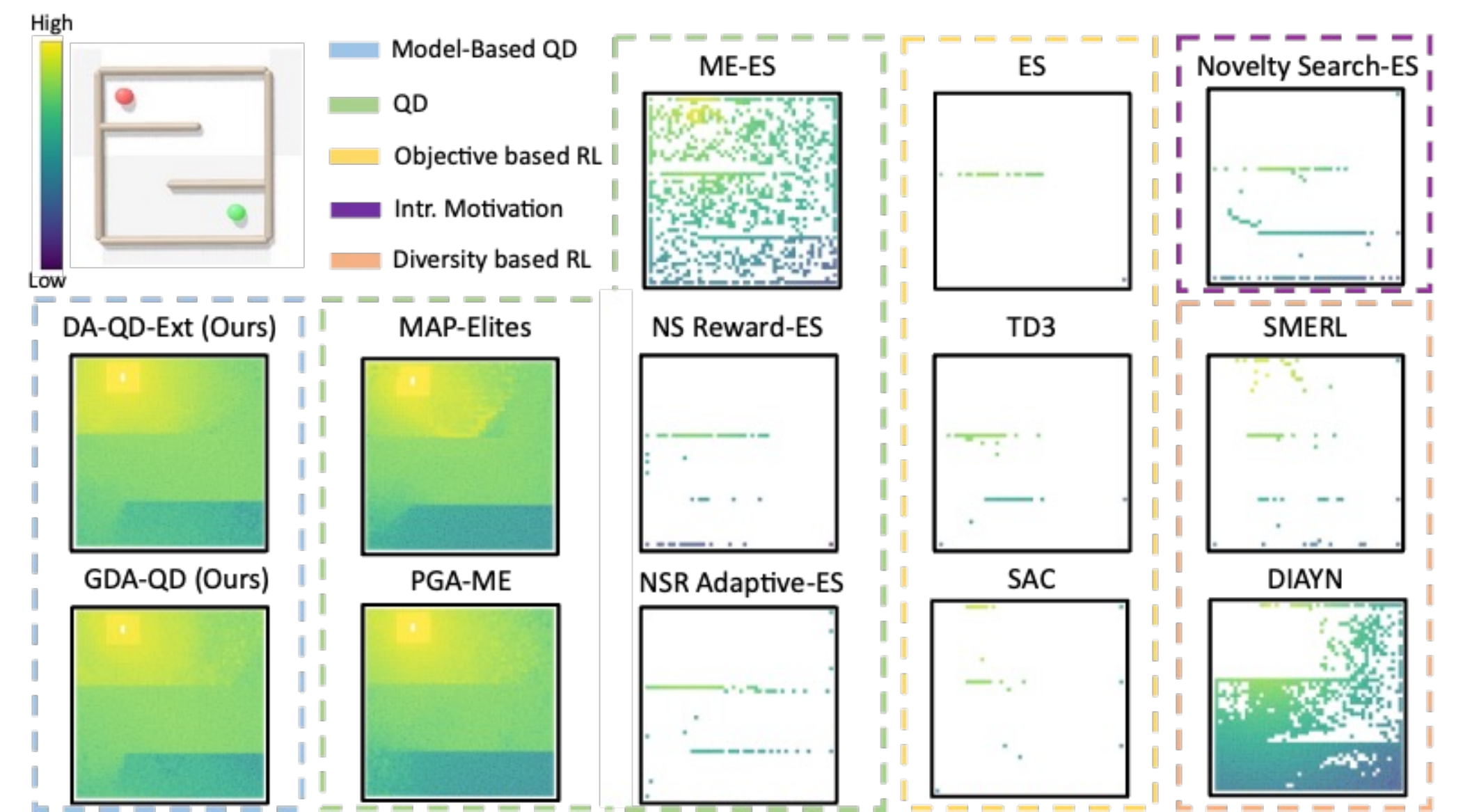}
    \caption{
        Final population of methods in the PointMaze environment (top left). 
        Each policy in the population is represented as a filled square in the final position (descriptor) it reaches within the episode.
        The color of the square of policy indicates the total reward, where the lighter the better.
    }
\label{fig:pointmaze_archives}
\end{figure*}

\begin{figure*}[!]
\centering
    \includegraphics[width = \textwidth]{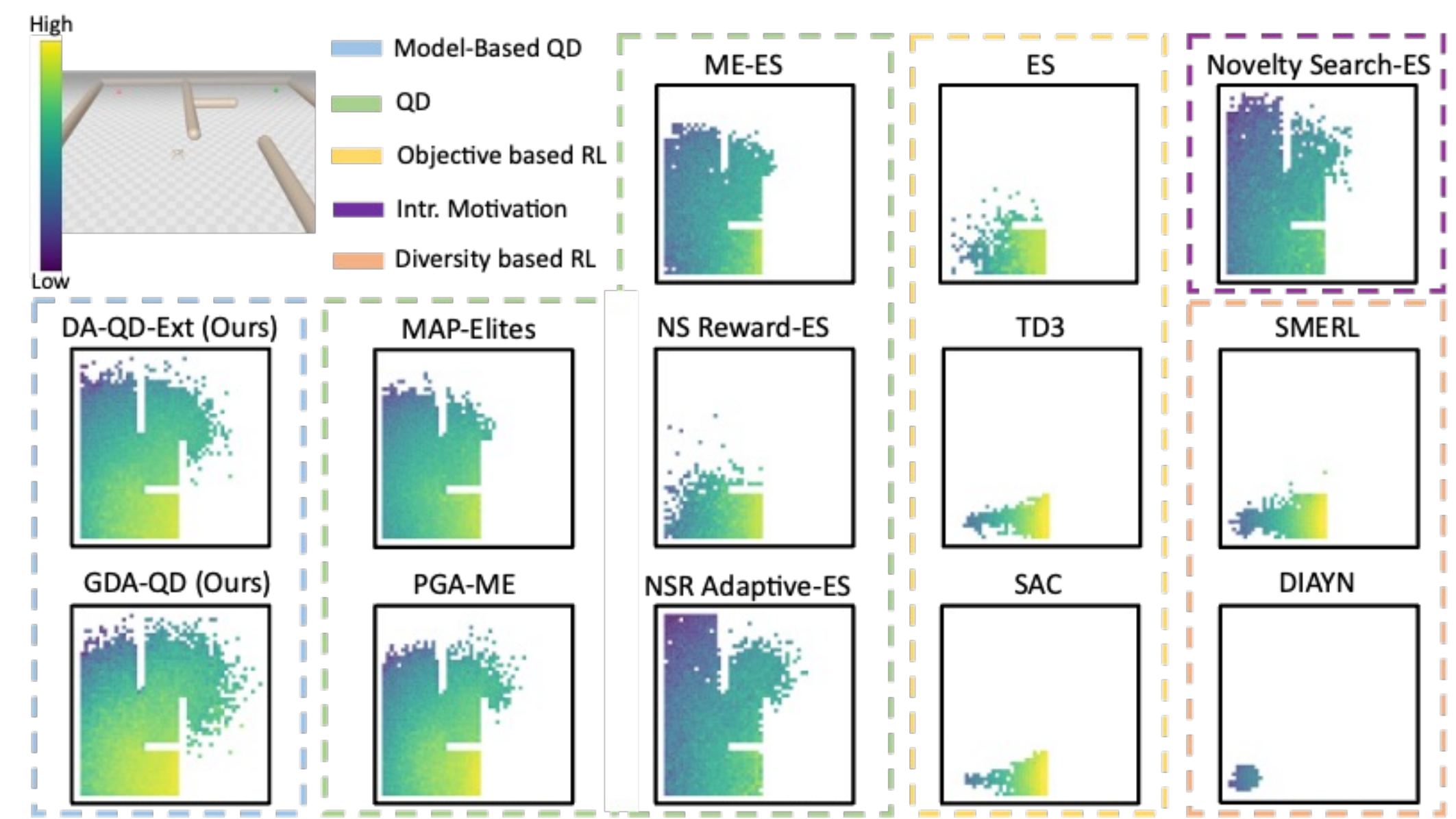}
    \caption{
        Final population of methods in the AntMaze environment (top left). 
        Each policy in the population is represented as a filled square in the final position (descriptor) it reaches within the episode.
        The color of the square of policy indicates the total reward, where the lighter the better.
        The boundaries of the maze clearly appears on this plot showing the deceptive nature of the task.
        Most objective-based baselines do not manage escape this.
    }
\label{fig:antmaze_archives}
\end{figure*}

\vspace{-2mm}
This section provides the plots of the population of policies for the PointMaze (Figure \ref{fig:pointmaze_archives}) and AntMaze (Figure \ref{fig:antmaze_archives}) environments respectively. 

Fig.~\ref{fig:graphs} plotted PointMaze for 500,000 evaluations for greater presentation clarity. We show the full performance curve after 1 million evaluations here to ensure convergence and the visualization of the final population of policies.

\begin{figure*}[b!]
\centering
    \includegraphics[width = 0.95\textwidth]{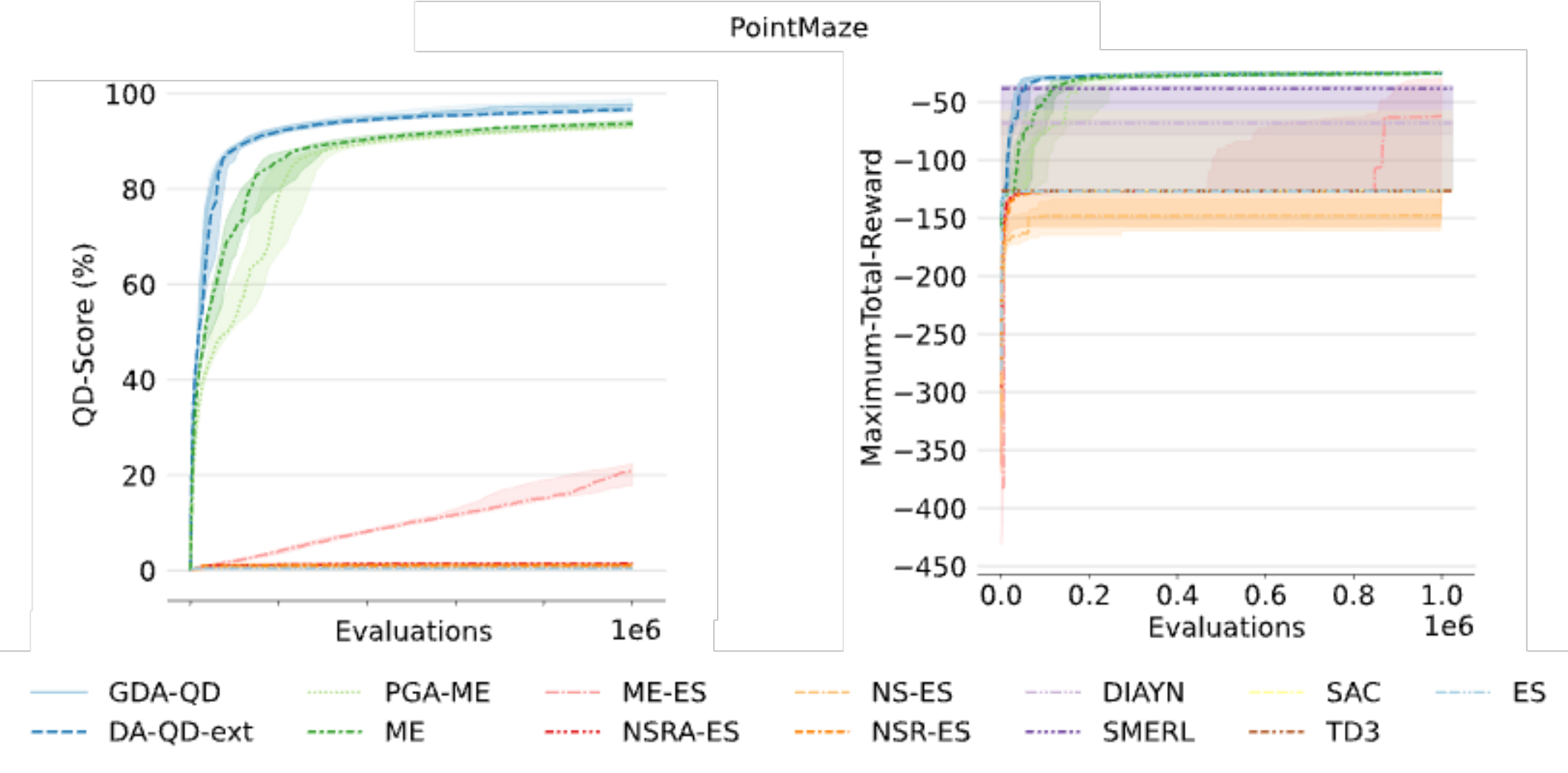}
    \caption{
        QD-Score (left) and Max-Total-Reward (right) on the PointMaze task.
        The solid line corresponds to the median over 15 replications and the shaded area to the first and third quartiles.
        This plot shows performance over 1 million evaluations.
    }
\label{fig:pointmaze1mil}
\end{figure*}


\newpage

\subsection{Coverage curves}
\begin{figure*}[h]
\centering
    \includegraphics[width = 0.99\textwidth]{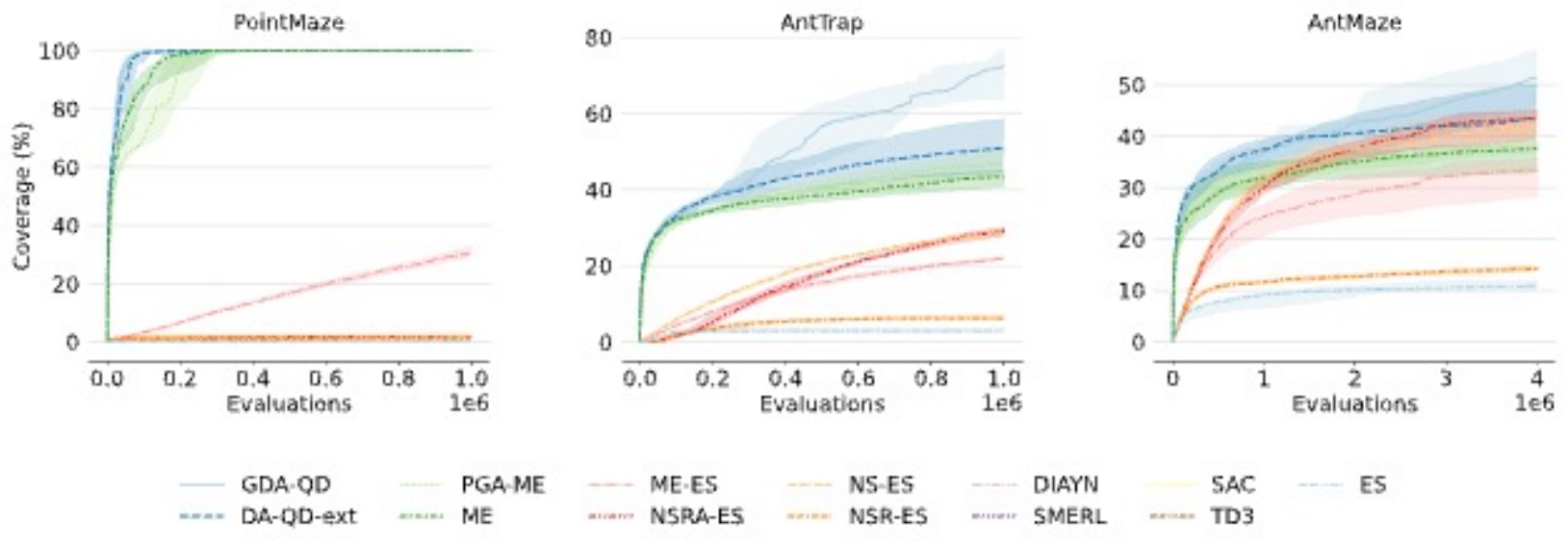}
    \caption{
        Coverage on the PointMaze, AntTrap and AntMaze tasks.
        Each experiment is replicated $15$ times, the solid line corresponds to the median over replications and the shaded area to the first and third quartiles.
    }
\label{fig:coverage}
\end{figure*}

Figure \ref{fig:coverage} shows the coverage curve for each of the baseline algorithms. This metric measures the diversity and supplements the Max. Total Reward (which measures the quality) and the QD-score (which measures both quality and diversity).

\subsection{Metrics Definitions and Discussion} \label{app:metrics}
As explained in Section~\ref{sec:exp}, we evaluate the performance of the algorithms on QD-Score, Max Total Reward and Coverage.
For fair comparison with single-policy and latent-conditioned RL baselines such as TD3, SAC, DIAYN and SMERL, we run the algorihtms for the same number of steps as evaluation budget $\times$ episode length.
This way, it has access to the same number of steps as the population based QD methods. 
We take the best performance of the agents throughout the learning process and plot them as a horizontal line to represent best performance.

For the exceptional case of the AntMaze, the reward at every timestep corresponds to the distance from the goal at every timestep.
This corresponds to a deceptive reward for this environment setup as discussed in~\citep{lehman2011novelty}.
To be fair to RL baselines, a time-step reward is given instead of a just a final reward at the end of episode which corresponds to the distance to the goal. 
However, this poses some problems during evaluation as the max total reward if not representative of solving task.
Solving the maze requires moving further away from the goal, resulting in decreasing rewards at every timestep and a final episode return that could possibly be lower that not solving the maze.
For example, moving quickly and getting stuck hovering around the region close to the wall near the goal would be able to give a high episodic return while definitely not solving the Maze.
This is evident and is done by the ES, TD3, SAC, SMERL baselines as seen in Figure~\ref{fig:antmaze_archives}.
One way to solve this could be to give a very large reward for reaching the goal. To avoid reward tuning, we opt to keep the simple reward function of distance to the goal.
Hence, using this metric naively would be evaluating the task wrongly.
Instead, we evaluate by normalising the max total reward obtained by the final distance travelled by the policy at the end of the episode. 
%Instead, we evaluate by normalising the max total reward obtained by the final distance of the policy at the end of the episode to the goal.

\section{NSRA-ES dynamics in hard exploration deceptive reward tasks}

\begin{figure*}[h]
\centering
    \includegraphics[width = \textwidth]{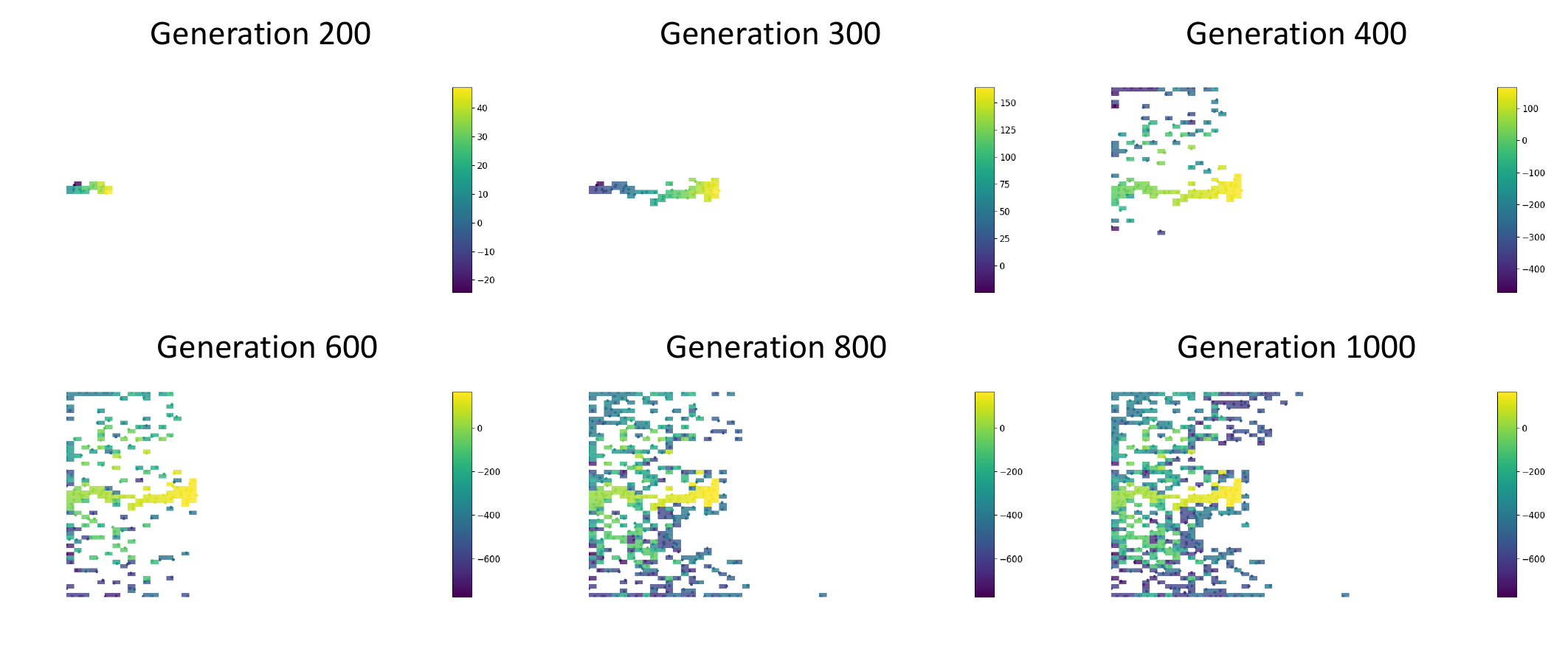}
    \caption{
        Population of NSRA-ES \cite{conti2018improving} at multiple generation number on the AntTrap environment. 
        Each policy in the population is represented with a dot in the final position it manages to reach within the episode.
        The color of the square around each policy indicates its total reward, the lighter the better.
        The obstacle clearly appears on this plot, as the empty area in the middle.
    }
\label{fig:nsra_es}
\end{figure*}

Figure~\ref{fig:nsra_es} shows a visualisation of the population generated by NSRA-ES at multiple generations number in AntTrap. 
NSRA-ES has an adaptive mechanism to weigh the exploitation and exploration in the objective given to its ES process. 
It starts with full exploitation, as can be seen in the first two visualisations in Figure~\ref{fig:nsra_es}. 
However, after a few hundred generations ($\sim 300$), NSRA-ES reaches the end of the trap, and thus the maximum total-reward it can expect to get in this direction. 
Thus, the adaptive mechanism kick in and the proportion of exploration raises, giving more weight to the novelty reward, as can clearly be seen in the following visualisations. 
However, NSRA-ES adaptive mechanism has one limitation: the proportion of exploitation in the ES-objective can only go up again when the algorithm finds solutions at least as performing as the best solutions found so far. However, in this complex control task, finding such high-performing solutions requires fine-tuning newly discovered solutions. Thus, after approximately $400$ generations, NSRA-ES remains in full exploration (novelty reward only) and does not manage to find high-performing solutions.

% Optionally include extra information (complete proofs, additional experiments and plots) in the appendix.
% This section will often be part of the supplemental material.

\end{document}